\documentclass[conference]{IEEEtran}
\usepackage{cite}
\usepackage{amsmath,amssymb,amsfonts}
\usepackage{algorithmic}
\usepackage[hidelinks]{hyperref}
\usepackage{listings}
\usepackage{graphicx}
\usepackage{textcomp}
\usepackage{xcolor}
\usepackage{tabularx}
\usepackage{booktabs}
\usepackage{tikz}
\usepackage{makecell}
\usepackage{tabularx}
\usepackage{svg}
\usepackage{gensymb}
\usepackage{subcaption}
\usepackage{algorithm2e}
\usepackage{xspace}
\usepackage[english]{babel}
\usepackage[numbers]{natbib}
\usepackage[nolist]{acronym}
\RestyleAlgo{ruled}
\usetikzlibrary{shapes,snakes, patterns}
\usetikzlibrary{calc} 
\usetikzlibrary{arrows.meta}
\usetikzlibrary{positioning}
\definecolor{olivegreen}{rgb}{0.33, 0.42, 0.18}
\lstdefinelanguage{sparql}{
morecomment=[l][\color{olivegreen}]{\#},
morestring=[b][\color{blue}]\",
morekeywords={SELECT,CONSTRUCT,DESCRIBE,ASK,WHERE,FROM,NAMED,PREFIX,BASE,OPTIONAL,FILTER,GRAPH,LIMIT,OFFSET,SERVICE,UNION,EXISTS,NOT,BINDINGS,MINUS,a},
sensitive=true
}
\definecolor{darkorange}{RGB}{170,70,0}
\definecolor{darkblue}{RGB}{0,60,130}
\tikzset{
	>=Latex,
	line/.style={draw,->},
	anode/.style={rectangle,draw,
		align=center,rounded corners,minimum height=4em,font=\strut},
	bnode/.style={anode,fill=white, font=\strut},
	cnode/.style={anode, fill=cyan!20, font=\strut},
treenode/.style = {circle,
	draw=black,thick, fill=white, align=center, minimum size=1cm},
root/.style     = {treenode, font=\footnotesize},
env/.style      = {treenode, font=\footnotesize}, 
dummy/.style    = {circle,draw}
	
}

\newcommand*{\figref}[1]{Fig.~\ref{#1}}
\def\BibTeX{{\rm B\kern-.05em{\sc i\kern-.025em b}\kern-.08em
    T\kern-.1667em\lower.7ex\hbox{E}\kern-.125emX}}

\definecolor{light-gray}{gray}{0.95}
\lstdefinestyle{mystyle}{
    backgroundcolor=\color{light-gray},  
    basicstyle=\ttfamily \footnotesize,
    breaklines=true,
    keywordstyle=\bfseries,
    morekeywords={SELECT, WHERE}
}

\newcommand{\onto}[1]{\texttt{#1}} % Für Elemente aus dem Graphen, wie ParX:expectsUnit
\newcommand{\fpb}{VDI/VDE~3682\xspace} %Einheitliche Benennung der 3682. "/VDE" rausstreichen, wenn nur VDI 3682 da stehen soll.

\begin{document}
\bstctlcite{IEEEexample:BSTcontrol}
\begin{acronym}
  \acro{odp}[ODP]{\textit{Ontology Design Pattern}}
  \acroplural{odp}[ODPs]{\textit{Ontology Design Patterns}}
  
  \acro{kbs}[KBS]{\textit{knowledge-based system}}
  \acroplural{kbs}[KBS]{\textit{knowledge-based systems}}
  
  \acro{sparql}[SPARQL] {\textit{SPARQL Protocol and RDF Query Language}}
  
  \acro{swrl}[SWRL] {\textit{Semantic Web Rule Language}}

  \acro{pge}[PGE]{\textit{Product Generation Engineering}}

  \acro{kpi}[KPI]{\textit{key performance indicator}}
  \acroplural{kpi}[KPIs]{\textit{key performance indicators}}

  \acro{cps}[CPS]{\textit{Cyber-Physical System}}
  \acroplural{cps}[CPS]{\textit{Cyber-Physical Systems}}

  \acro{owl}[OWL]{\textit{Web Ontology Language}}
  \acro{rdf}[RDF]{\textit{Resource Description Framework}}
  \acro{fpd}[FPD]{\textit{Formalized Process Description}}
  \acro{rtm}[RTM]{\textit{Resin Transfer Molding}}
  
\end{acronym}

\title{Consistency Verification in Ontology-Based Process Models with Parameter Interdependencies} % Muss besser an inhalt angepasst werden

\author{
\IEEEauthorblockN{
Tom Jeleniewski\IEEEauthorrefmark{1}, 
Hamied Nabizada\IEEEauthorrefmark{1}, 
Jonathan Reif\IEEEauthorrefmark{1},
Felix Gehlhoff\IEEEauthorrefmark{1},
Alexander Fay\IEEEauthorrefmark{2}}\\
\IEEEauthorblockA{\IEEEauthorrefmark{1}Institute of Automation Technology\\
Helmut Schmidt University, Hamburg, Germany\\
Email: \{tom.jeleniewski, hamied.nabizada, jonathan.reif, felix.gehlhoff\}@hsu-hh.de} \\
\IEEEauthorblockA{\IEEEauthorrefmark{2}Chair of Automation \\ Ruhr University Bochum, Bochum, Germany\\
Email: alexander.fay@rub.de}}

\maketitle
\begin{abstract}
The formalization of process knowledge using ontologies enables consistent modeling of parameter interdependencies in manufacturing. 
These interdependencies are typically represented as mathematical expressions that define relations between process parameters, supporting tasks such as calculation, validation, and simulation. 
To support cross-context application and knowledge reuse, such expressions are often defined in a generic form and applied across multiple process contexts. 
This highlights the necessity of a consistent and semantically coherent model to ensure the correctness of data retrieval and interpretation. 
Consequently, dedicated mechanisms are required to address key challenges such as selecting context-relevant data, ensuring unit compatibility between variables and data elements, and verifying the completeness of input data required for evaluating mathematical expressions. 
This paper presents a set of verification mechanisms for a previously developed ontology-based process model that integrates standardized process semantics, data element definitions, and formal mathematical constructs. 
The approach includes (i) SPARQL-based filtering to retrieve process-relevant data, (ii) a unit consistency check based on expected-unit annotations and semantic classification, and (iii) a data completeness check to validate the evaluability of interdependencies. 
The applicability of the approach is demonstrated with a use case from Resin Transfer Molding (RTM), supporting the development of machine-interpretable and verifiable engineering models.
\end{abstract}


\section{Introduction} \label{sec:Introduction}
The ongoing transformation in manufacturing is characterized by a growing demand for individualized products, shorter product life cycles, and decreasing lot sizes~\cite{Jarvenpaa.2016}. 
To remain competitive in this dynamic environment, modern production systems must be highly flexible, reconfigurable, and capable of adapting to changing product requirements and operational constraints \cite{Afazov.2013}. 
As a result, the complexity of production processes and their configurations increases significantly~\cite{Luder.2017}. 

In response to these challenges, data-driven engineering practices and semantic technologies have become key enablers for the design, optimization, and validation of flexible production systems. 
Particularly in the context of \acp{cps}, machine-interpretable and semantically precise models are essential. 
\textit{Semantic Web} technologies offer a promising foundation and are increasingly used  for such tasks~\cite{Sabou.2020}. 
They allow the formalization of complex relations between system components, data semantics, and process behavior in a way that is both machine-readable and compliant with industrial standards.

However, support for mathematical calculations, which is often required in engineering domains, remains a significant challenge for Semantic Web technologies~\cite{Sabou.2020}. 
Nonetheless, initial approaches have been proposed to address this issue, such as the introduction of \textit{OpenMath-RDF} by \citet{wenzel2021openmath}.

Our previous works \cite{Jeleniewski.SemanticModel2023,Jeleniewski.2024} have shown how parameter interdependencies in manufacturing processes can be explicitly modeled using such technologies, integrating and aligning process semantics (e.g., \fpb~\cite{VDIVDEGesellschaftMessundAutomatisierungstechnik.05.2015}), data element ontologies (e.g., DIN EN 61360~\cite{DIN.Juni2018}), and formal mathematical expressions using \textit{OpenMath-RDF}~\cite{wenzel2021openmath}. 
These mathematical interdependencies are used to express physical laws affecting production processes, domain-specific engineering constraints, or \acp{kpi}. 

Their explicit modeling supports tasks such as process simulation, model checking, and parameter estimation via calculations. 
However, this increased expressiveness also introduces challenges. 
Since the equations are usually modeled in a generic and reusable way, a variable can be linked to multiple data elements, although not all of them are relevant in every specific process context. 
Additionally, units may be incorrectly assigned, or input variables may not have associated data, either because the data has not been modeled or because it has not been connected to the variable. %old:or input variables may be insufficiently grounded in data. 
Ensuring that such interdependencies can be reliably applied across different process contexts is essential to support their reusability in model-based engineering workflows. 

For these reasons, consistency verification mechanisms are necessary to ensure the quality of the modeled knowledge. 
Therefore, this paper addresses the following questions: (i) how context-relevant data elements can be identified, (ii) how to verify whether the units provided by data elements match the expected units of a formula and (iii) how to check whether a formula has access to all necessary input information for its variables. %old: evaluation. 
 Semantic Web technologies and the associated checking methods (e.g., \ac{sparql}) are suitable for this purpose \cite{Sabou.2020} and are therefore applied in this contribution.

The structure of this paper is as follows: Section~\ref{sec:RelatedWork} provides an overview of existing approaches for semantic modeling of process knowledge and interdependencies in the manufacturing domain. Section~\ref{sec:Approach} introduces the proposed verification mechanisms, including \ac{sparql}-based filtering, unit consistency checking, and data availability verification. The applicability of these mechanisms is demonstrated in Section~\ref{sec:Application} using an exemplary scenario from a \ac{rtm} process. Section~\ref{sec:conclusion} summarizes the results and outlines future research directions.

\section{Related Research}  
\label{sec:RelatedWork}
As production processes are frequently reconfigured or redesigned, a key challenge lies in capturing and formalizing the complex interdependencies among process parameters. Traditional information models often fall short in this regard, as they typically focus on structural or entity-based representations without integrating causal or functional relations between parameters.
Initial contributions to parameter modeling focus on simulation and qualitative representations. \citet{Denkena.2011Simulationbaseddimensioningof} and \citet{Grigoriev.2013} highlight the importance of dimensioning and optimization based on physical principles and empirical data, but lack a semantic structure that supports reuse and automated reasoning.

\citet{Hoang.2017} propose a model for describing interdependencies between product, process, and resource parameters to support mechatronic system adaptation. Although interdependencies are acknowledged, the approach remains qualitative, limiting its applicability in precise parameter prediction.

Semantic Web technologies offer promising tools to formalize domain knowledge in a reusable and machine-interpretable format. Ontology-based approaches have therefore gained attention in manufacturing. 
An example is \citet{Liang.2018}, who introduces an ontology for process planning in the additive layer manufacturing domain, as well as \citet{Cheng.92016}, who developed ontologies for \textit{Industry 4.0} demonstration production lines. Both examples, however, are tailored to specific applications and do not integrate industrial standards, thereby limiting their generalizability.
The approach presented by \citet{Hildebrandt.2020.Ontology}, in contrast, encompasses methodologies for creating domain-specific ontologies in the context of \acp{cps} by developing and using so-called \acp{odp}, which contain standard-based and compliant concepts and terminology.
This approach formed the foundation for several works like \citet{KHV+_AFormalCapabilityand_2020} for describing machine capabilities and skills or \citet{Gill.2023Utilisationofsemantic} for describing maintenance processes for aircraft components.

The need to express functional interdependencies in process models was addressed in our previous work \cite{Jeleniewski.SemanticModel2023}, where the alignment ontology \textit{ParX}\footnote{\url{https://github.com/hsu-aut/ParX}} integrating multiple \acp{odp} to ensure semantic consistency, reusability, and extensibility across manufacturing domains was introduced. The \fpb serves as the structural backbone, defining a process operator as a transformation of inputs (information, energy, or product) into outputs by a technical resource. Its associated \ac{odp}\footnote{\url{https://github.com/hsu-aut/IndustrialStandard-ODP-VDI3682}} formalizes the semantics and states of process operators, supporting the modeling of sequential and nested steps. To enable a specification for structures of technical resources the VDI 2206 \ac{odp}\footnote{\url{https://github.com/hsu-aut/IndustrialStandard-ODP-VDI2206}} is aligned with the semantic model. 

Parameter representation relies on the DIN EN 61360 \ac{odp}\footnote{\url{https://github.com/hsu-aut/IndustrialStandard-ODP-DINEN61360}}, which separates instance and type descriptions for clearer and reusable data semantics. To standardize units, a UNECE-derived \ac{odp}\footnote{\url{https://github.com/hsu-aut/IndustrialStandard-ODP-UNECE-UoM}} links unit codes and symbols according to \cite{UNECE.2010} to these type descriptions, ensuring system-wide compatibility.

To express interdependencies as mathematical relations, suitable formalisms have been integrated into the alignment ontology. Building on prior work \cite{Marchiori.2003, Lange.2013}, \citet{wenzel2021openmath} introduced \textit{OpenMath-RDF} as a representation of mathematical expressions in knowledge graphs. This provides the foundation for capturing parameter interdependencies. A systematic method to integrate these interdependencies into formalized process models was introduced in \cite{Jeleniewski.IntegratingInterdependenciesin2023}.

As shown in \cite{Jeleniewski.2024}, this approach enables both the composition and decomposition of processes together with their associated interdependencies. 
To handle the resulting complexity across multiple levels of abstraction, filtering mechanisms were introduced to extract only context-relevant process data. This, in turn, revealed the need for dedicated consistency checks, which form the focus of the present work.
The verification mechanisms proposed here are specifically tailored to the structural characteristics of the \textit{ParX} model and are intended to assist engineers in the reliable application of reusable interdependencies within knowledge-based reasoning.

\section{Consistency Verification Approach} \label{sec:Approach}
In knowledge-based modeling environments, particularly those supporting design and configuration of \ac{cps}, consistency verification plays a central role~\cite{Sabou.2020}. 
As the amount and complexity of interconnected data increases in the semantic knowledge graph introduced in our previous work~\cite{Jeleniewski.SemanticModel2023}, so does the risk of modeling inconsistencies. These include, for example, missing data bindings, unit mismatches, and incomplete definitions of parameter interdependencies. To address these challenges, this section introduces a set of verification techniques targeting typical sources of semantic inconsistencies in such a knowledge graph. 
 
First, a \ac{sparql}-based filtering mechanism is introduced in Section~\ref{subsec:relevantC}, ensuring that only those data elements relevant to the considered process and its associated interdependence expressions are used. 
This is followed by a unit consistency check in Section~\ref{subsec:unitC}, which verifies that each mathematical variable is semantically compatible with the unit of its connected data element. 
Finally, in Section~\ref{subsec:ComputeC}, a data availability check is performed to ensure that target variables of a mathematical interdependency have assigned data elements required for a calculation based on the knowledge graph. 

\subsection{SPARQL-Based Filtering of Context-Relevant-Data} \label{subsec:relevantC}
In order to perform consistency verification and calculations based on described interdependency relations, it is necessary to retrieve only those data elements from the knowledge graph that are contextually relevant to a specific process and the associated mathematical interdependency. 
The underlying challenge lies in the fact that the same formula (e.g., a physical law) can be used in multiple contexts (e.g. see \eqref{eq:filltime} in Section \ref{sec:Application}), resulting in several data elements referencing the same mathematical variable. 

For example, a variable such as \( Q \) (flow rate) may be reused in different process steps such as filling or draining, each connected to a different data element in the knowledge graph. 
Without filtering for context, a formula might mistakenly use data from an unrelated part of the process. 
To ensure that the correct data element is selected for a variable within an interdependency description, a dedicated \ac{sparql}-based filtering mechanism was developed, as introduced by \citet{Jeleniewski.2024}. 
The filtering query operates by navigating the semantic structure of the process model. 
Starting from a selected process operator, the query recursively collects all associated states, technical resources, and any data elements connected to any of these elements (including the selected process operator). %old: all associated states, technical resources and their connected data elements. 
The query also ensures that these data elements are connected to the variables referenced in the respective mathematical expression describing the interdependency of the process operator. 
Consequently, only those data elements are identified and extracted from the knowledge graph that are required for the evaluation of a specific interdependency.
This ensures that no incorrect data from other processes or resources is used for calculations based on the interdependency description.

In the \ac{sparql}-based filtering mechanism introduced in~\cite{Jeleniewski.2024}, several \texttt{FILTER EXISTS} clauses are used to isolate contextually relevant data elements associated with or related to a particular process. 
Listing~\ref{lst:ConsistencyCheck} shows a refactored version of this \ac{sparql}-based filtering mechanism, in which the process only needs to be specified once. 
This refactoring improves clarity and makes the query easier to reuse across different scenarios.

\begin{lstlisting}[caption={SPARQL filter function pattern to get only data used in the considered process based on \cite{Jeleniewski.2024}}, language=sparql, keywordstyle=\color{blue}\bfseries, label={lst:ConsistencyCheck}]
FILTER (
EXISTS {?proc VDI3682:hasOutput ?output.
        ?output DINEN61360:has_Data_Element ?data.}
||EXISTS {?proc VDI3682:hasInput ?input.
        ?input DINEN61360:has_Data_Element ?data.}
||EXISTS {?proc VDI3682:isAssignedTo ?tr.
        ?tr DINEN61360:has_Data_Element ?data.}
||EXISTS {?proc  DINEN61360:has_Data_Element ?data.}
)
\end{lstlisting}

\subsection{Unit Consistency Verification} \label{subsec:unitC}

A critical semantic inconsistency arises when the unit of the provided data mismatches the expected unit of a mathematical variable of an interdependency description. 
For example, a formula may expect the diameter of a component in millimeters to calculate the required rotational speed. 
If the connected data element provides the diameter in inches and no unit conversion is performed, the resulting speed calculation will be incorrect.

These types of mismatches can occur particularly when integrating data from suppliers or heterogeneous systems using different unit conventions, or when the process model is created manually without technical assistance.
If such inconsistencies are not detected at an early stage, they can lead to incorrect parameter evaluations or misinformed engineering decisions.

This highlights the need for a verification approach specifically targeting modeling errors caused by unit mismatches between units of data elements and the expected units of mathematical variables. 
In our previous work~\cite{Jeleniewski.SemanticModel2023, Jeleniewski.2024}, we developed a knowledge graph in which the variables of mathematical equations are integrated into the process description and connected to data elements that include unit information.
The units of data elements are modeled via semantic classification of their type description according to the DIN EN 61360 standard.
Instead of relying on potentially error-prone string comparisons for unit matching, the type descriptions are semantically classified as subclasses of the corresponding \texttt{UNECE:Unit}, providing a robust basis for unit consistency verification. 
However, the knowledge graph does not yet support the explicit specification of the unit a formula expects for a given variable, which makes it impossible to verify whether the unit provided by a connected data element is semantically consistent with the formula’s intended interpretation. 

\begin{figure*} [h]
    \centering
    \includegraphics[width=0.8\linewidth]{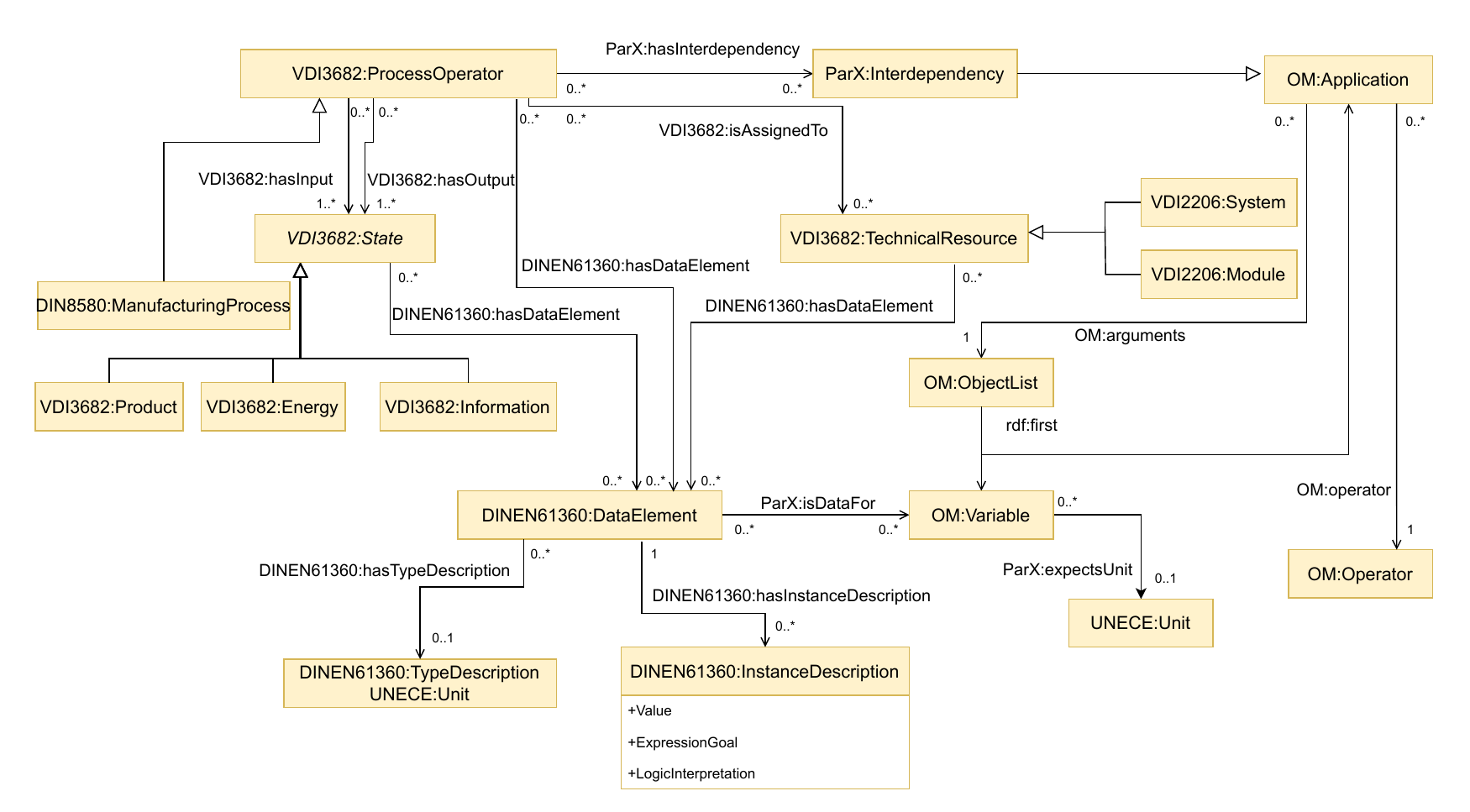}
    \caption{Class diagram of the extended alignment ontology based on \cite{Jeleniewski.SemanticModel2023}}
    \label{fig:uml}
\end{figure*}

To address this, we extend our knowledge graph by introducing the object property \onto{ParX:expectsUnit}, which explicitly relates mathematical variables~(\onto{OM:Variable})  with their semantically expected units from the UNECE unit of measurement ontology (\onto{UNECE:Unit}). 
The extended class diagram is shown in \figref{fig:uml}. 

The association between variables and expected units can be expressed formally, indicating that every variable is expected to reference a unit from the UNECE ontology via \onto{ParX:expectsUnit}~(see \eqref{eq:unitStructuralAxiom}). 

\begin{equation}\small
\texttt{OM:Variable} \sqsubseteq \exists\, \texttt{ParX:expectsUnit}.\texttt{UNECE:Unit}
\label{eq:unitStructuralAxiom}
\end{equation}

Consequently, unit consistency verification can be performed using the \ac{sparql} query shown in Listing~\ref{lst:UnitConsistencyCheck}.
This query identifies all mathematical variables whose assigned data elements (\onto{ParX:isDataFor}) have units that do not match the declared expected units (\onto{ParX:expectsUnit}). 
By exposing such mismatches automatically, the mechanism ensures that variable-unit assignments in the knowledge graph remain consistent and technically valid.

\begin{lstlisting}[caption={Unit consistency verification using SPARQL}, language=sparql, keywordstyle=\color{blue}\bfseries, label={lst:UnitConsistencyCheck}]
PREFIX ParX: <http://www.hsu-hh.de/aut/ParX#>
PREFIX DINEN61360: <http://www.w3id.org/hsu-aut/DINEN61360#>

SELECT ?variable ?expectedUnit ?actualType ?actualUnit
WHERE {
  ?variable ParX:expectsUnit ?expectedUnit .
  ?dataElement ParX:isDataFor ?variable ;
               DINEN61360:has_Type_Description ?actualType .
  ?actualType a ?actualUnit .

  FILTER(?actualUnit != ?expectedUnit)
}
\end{lstlisting}

\subsection{Ensuring data availability for interdependency effect calculation} \label{subsec:ComputeC}
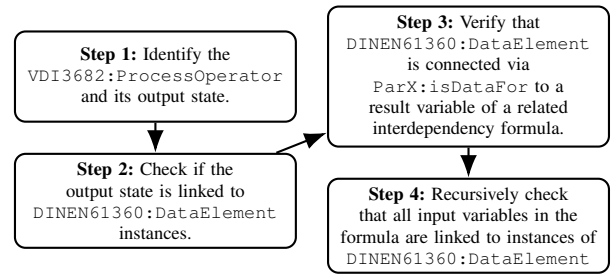
\begin{figure}[h]
    \centering
\begin{tikzpicture}[
  node distance=0.4cm and 0.4cm,
  every node/.style={font=\scriptsize},
  process/.style={rectangle, rounded corners, draw=black, thick, minimum width=3cm, minimum height=1.2cm, text width=3.5cm, align=center},
  arrow/.style={-{Latex[length=3mm]}, thick}
]

% Nodes
\node[process] (step1) {\textbf{Step 1:} Identify the \texttt{VDI3682:ProcessOperator} and its output state.};
\node[process, below=of step1] (step2) {\textbf{Step 2:} Check if the output state is linked to \texttt{DINEN61360:DataElement} instances.};
\node[process, right=of step1] (step3) {\textbf{Step 3:} Verify that \texttt{DINEN61360:DataElement} is connected via \texttt{ParX:isDataFor} to a result variable of a related interdependency formula.};
\node[process, below=of step3] (step4) {\textbf{Step 4:} Recursively check that all input variables in the formula are linked to instances of \texttt{DINEN61360:DataElement}};

% Arrows
\draw[arrow] (step1) -- (step2);
\draw[arrow] (step2) -- (step3);
\draw[arrow] (step3) -- (step4);

\end{tikzpicture}
  \caption{Workflow for verifying the availability of input data for computing process outputs based on modeled interdependencies}
    \label{fig:ComputabilityCheckworkflow}
\end{figure}

Process models often define outputs that represent system state variables, such as product characteristics or \acp{kpi}, derived through mathematical relations involving multiple input parameters.  
Ensuring that these outputs can be computed based on the available data is a challenging task, particularly when input data is missing or data bindings are incomplete.
To address the risk of incomplete data inputs, we propose a consistency verification method explicitly focused on process outputs derived from process operators defined according to \fpb. 
Each process operator~(\onto{VDI3682:ProcessOperator}) generates specific outputs such as products and information represented as states (\onto{VDI3682:State}). 
According to the DIN EN 61360 semantic modeling approach, these states are connected via~(\onto{DINEN61360:hasDataElement}) to concrete data representations (\onto{DINEN61360:DataElement}). 
These data elements serve as result variables in mathematical interdependencies modeled using \textit{OpenMath-RDF} (\onto{ParX:isDataFor}).
In \figref{fig:ComputabilityCheckworkflow}, the verification workflow is illustrated. 
 
In \textbf{Step~1}, the verification begins by identifying a process operator (\onto{VDI3682:ProcessOperator}) and its associated output states (\onto{VDI3682:State}). This defines the specific process context in which a calculation is to be performed. 

\textbf{Step~2} ensures that the output states are connected to a corresponding \onto{DINEN61360:DataElement}, which serves as the formal data representation of the process result. 
This relation ensures that an output state has a concrete data representation in the model, which is a prerequisite for associating it with a result variable in an interdependency formula.

\textbf{Step~3} checks whether the data elements are connected to a result variable using the \onto{ParX:isDataFor} property, indicating that it represents the outcome of a modeled mathematical interdependency. 

The most critical part of the verification is performed in \textbf{Step~4}, which involves a recursive consistency check of the formula. 
This step verifies that all input variables used within the expression are themselves connected to data elements. 
If a formula contains nested operations or intermediate variables defined through other interdependencies, the verification is applied recursively along the entire dependency chain. 
This ensures that the complete mathematical expression is grounded in available and connected data. 

The whole verification method can be automated through a single, reusable \ac{sparql} query, which is shown in Listing \ref{lst:ComputabilityyCheck}.
It returns processes and their variables which are not connected to a data element providing data for calculations.
Please note that the context filter pattern presented in \ref{subsec:relevantC} must be inserted at the commented position in the query.

\begin{lstlisting}[caption={Data availability verification using SPARQL }, language=sparql, keywordstyle=\color{blue}\bfseries, label={lst:ComputabilityyCheck}]
PREFIX OM: <http://openmath.org/vocab/math#>
PREFIX ParX: <http://www.hsu-hh.de/aut/ParX#>
PREFIX VDI3682: <http://www.w3id.org/hsu-aut/VDI3682#>
PREFIX DINEN61360: <http://www.w3id.org/hsu-aut/DINEN61360#>
PREFIX RDF: <http://www.w3.org/1999/02/22-rdf-syntax-ns#>

SELECT DISTINCT ?process ?missingVar
WHERE {
  ?process a VDI3682:ProcessOperator ;
           VDI3682:hasOutput ?state ;
           ParX:hasInterdependency ?formula .
  ?state DINEN61360:has_Data_Element ?outputDE .
  ?outputDE ParX:isDataFor ?resultVar .

  FILTER EXISTS { ?formula om:arguments ?argList . }

  {
    ?formula OM:arguments ?argList .
    ?argList rdf:rest*/rdf:first ?argExpr .
    ?argExpr (OM:arguments/rdf:rest*/rdf:first)* ?missingVar .
    ?missingVar a OM:Variable .
  }
  UNION
  {
    ?formula OM:arguments ?argList .
    ?argList rdf:rest*/rdf:first ?missingVar .
    ?missingVar a OM:Variable .
  }

  FILTER NOT EXISTS {
    ?dataElement ParX:isDataFor ?missingVar .
   # {$CONTEXT FILTER PATTERN$}
  }
}

\end{lstlisting}

\section{Application} \label{sec:Application}

\ac{rtm} is an established manufacturing technique for producing high-quality fiber-reinforced plastic parts, especially in applications where weight savings and geometric complexity are critical, such as in the aerospace domain~\cite{Laurenzi.2012}.
\begin{figure}[h]
    \centering
    \includegraphics[width=0.8\columnwidth]{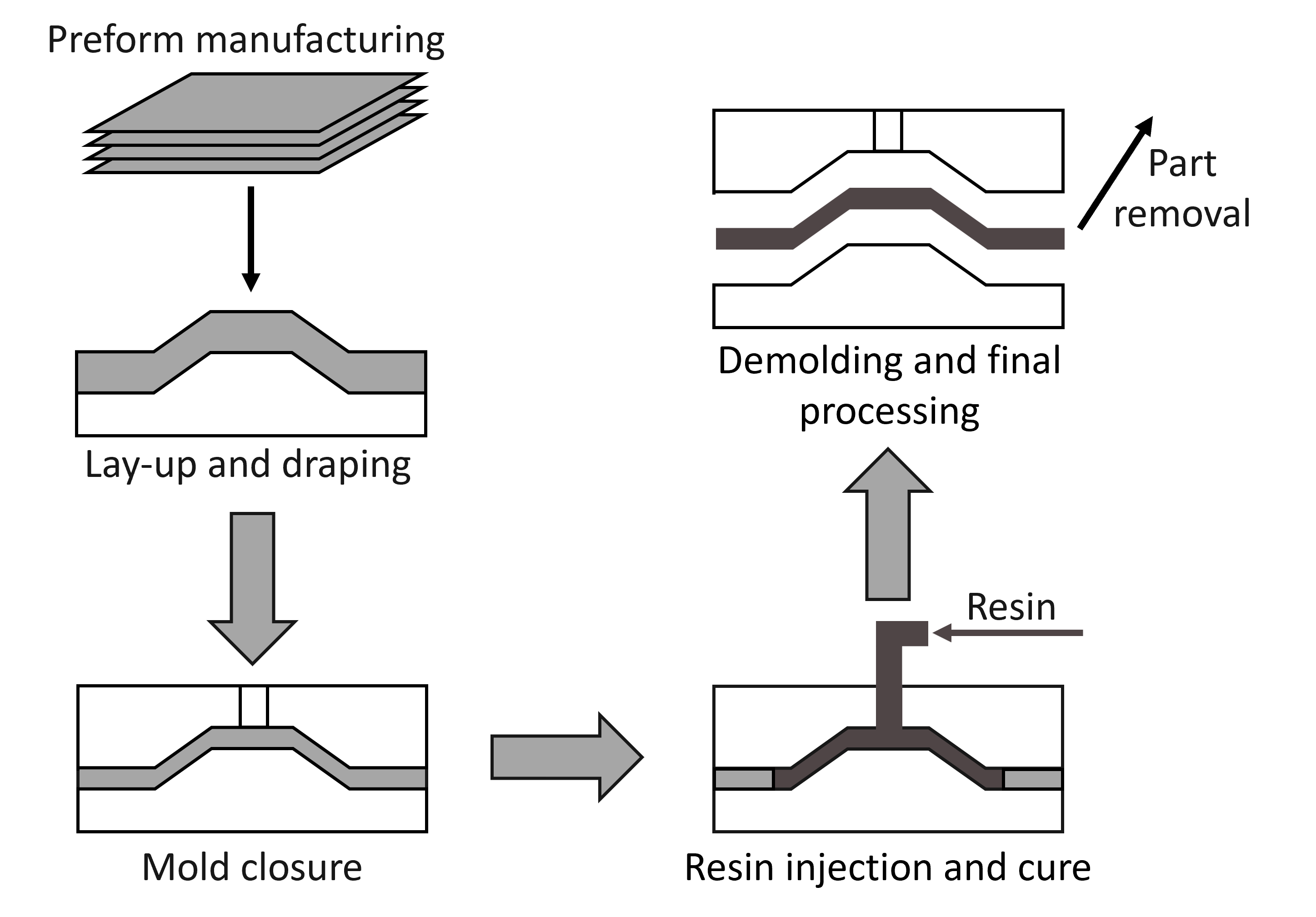}
    \caption{Schematic illustration of process steps in \ac{rtm} according to \cite{Advani.2002}}
    \label{fig:rtm}
\end{figure}
In \figref{fig:rtm}, process steps of an \ac{rtm} process are illustrated schematically.
The process begins with the placement of a dry fiber preform into a rigid mold, which is then closed. 
Liquid thermoset resin is injected under controlled pressure into the mold cavity supported by vacuum, where it infiltrates the fiber. 
After full saturation, the resin is activated by heat and undergoes a curing reaction.

\begin{figure}[]
    \centering
    \includegraphics[width=\columnwidth]{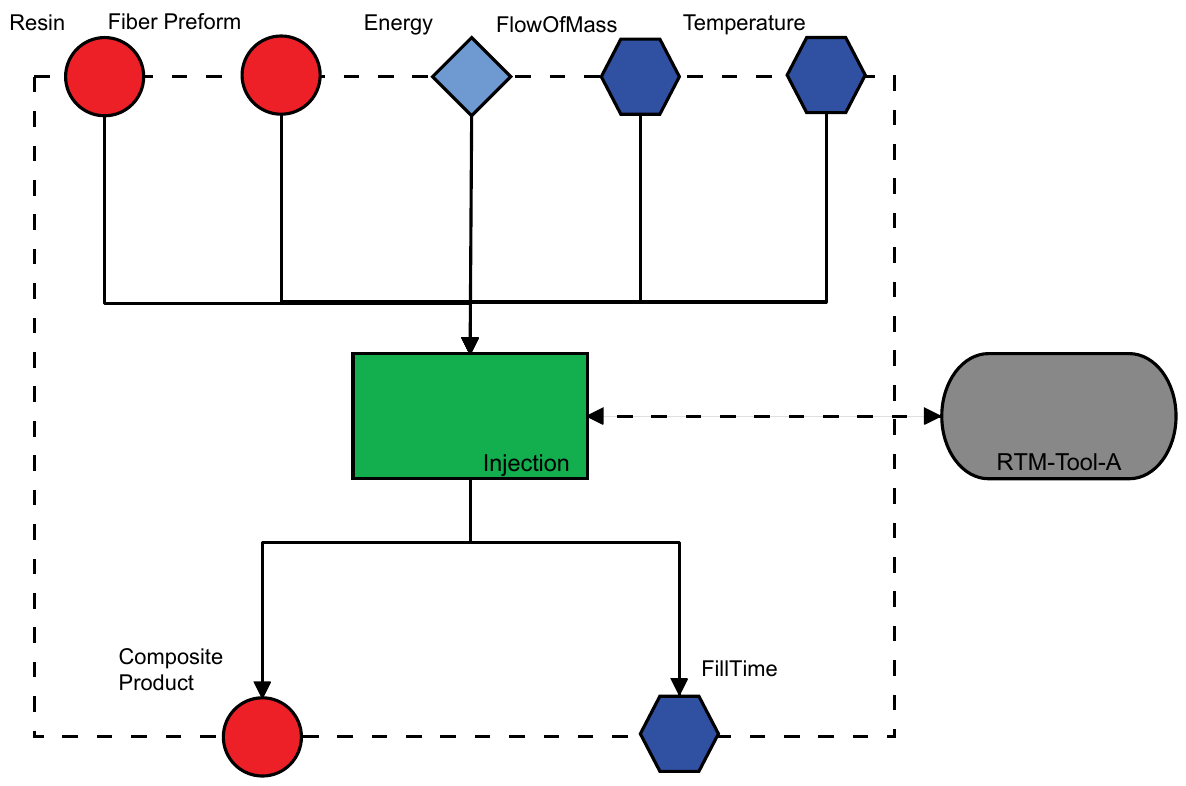}
    \caption{Formalized Process Description of an \ac{rtm}-Injection subprocess according to \fpb}
    \label{fig:injection}
\end{figure}

\ac{rtm} enables precise control over fiber alignment, part geometry, and surface quality, while also supporting repeatable and cost-efficient production~\cite{Laurenzi.2012}. 
The injection process in the form of a Formalized Process Description according to \fpb is shown in \figref{fig:injection}.
This figure provides an overview of products (red circle), energy (light blue diamond), and information (dark blue hexagon) entering a process operator or process step (green box) as inputs and being transformed into corresponding outputs. The assignment of a technical resource (grey) is also depicted.

The outcome of the \ac{rtm} process strongly depends on various interdependencies between process parameters, including injection pressure, resin viscosity, mold temperature, and cavity volume~\cite{Laurenzi.2012}. 
Capturing the required expert knowledge about processes, products, resources, and parameter interdependencies remains a challenge, as it is often not formally defined or available in a machine-readable format. 
Therefore, \ac{rtm} provides a suitable context for applying the proposed modeling approach and the corresponding consistency verification approach.

In this work, the process serves as an application scenario to demonstrate how semantic filtering, unit compatibility validation, and computability checking can support the reliable derivation of process-relevant output values and \acp{kpi} based on modeled interdependencies. 
As no standard-based semantic models for this process exist in the current literature, the model was created in close collaboration with an industrial partner to ensure domain relevance and practical applicability.

Accordingly, the individual process steps were modeled according to \fpb using the modeling tool described in \cite{Nabizada.2020}. 
The resulting JSON representation was then mapped into the graph structure of the \fpb \ac{odp}.\footnote{The mapper used for this is available at: \url{https://github.com/hsu-aut/fpb-owl-mapper}} 
\begin{figure}[h]
    \centering
    \resizebox{\columnwidth}{!}{
    % \scalebox{0.60}{
        \begin{tikzpicture}
        [env/.style={circle, draw=gray!60, fill=gray!10, thick, minimum size=10mm}]
         [envb/.style={circle, draw=gray!60, fill=orange!10, thick, minimum size=10mm}]
			\begin{scope}[
			every node/.style = {font=\large, thick}, 
			every label/.style = {font=\large, align=center}]
				\node[env, label=below:{\text{ex:CavityVolume-A} \\ \texttt{DIN61360:DataElement}},draw=gray] (CVA) at (-17,-3) {};
				\node[env, label={\text{ex:RTM-Tool-A} \\ \texttt{VDI3682:TechnicalResource}},draw=gray] (TRA) at (-17, -0.5) {};
				\node[env, label={\text{ex:Injection} \\ \texttt{VDI3682:ProcessOperator}}, draw=gray] (PO) at (-18,5) {};
                \node[env, label={\text{ex:FillTime} \\ \texttt{VDI3682:Information}}, draw=gray] (OI) at (-12,1) {};
                \node[env,pattern=crosshatch, label={\text{ex:t} \\ \texttt{OM:Variable}}, draw=gray] (t) at (-6.5,0) {};
                \node[env, label={\text{ex:TimeDE} \\ \texttt{DIN61360:DataElement}}, draw=gray] (TDE) at (-13,-2.2) {};
                \node[env,pattern=crosshatch, label={\text{ex:Time} \\ \texttt{OM:Application}}] (AppT) at (-13,6) {};
                \node[env,pattern=crosshatch, label={\text{ex:TimeFrac} \\ \texttt{OM:Application}}] (App) at (-5,3) {};
                \node[env,pattern=crosshatch, label={\text{CD:arith1/divide} \\ \texttt{OM:Object}}] (Op) at (-4,6.5) {};
                \node[env, pattern=crosshatch,label={\text{CD:relation1/eq} \\ \texttt{OM:Object}}] (OpEq) at (-11,9.5) {};
                \node[env,pattern=crosshatch, label={\text{ex:TimeAttribute2} \\ \texttt{rdf:List}}] (AL2) at (-1.7,3) {};
                \node[env, pattern=crosshatch, label=below:{\text{ex:TimeAttribute1} \\ \texttt{rdf:List}}] (AL1) at (-4,-1) {};
                \node[env, pattern=crosshatch, label=below:{\text{ex:EqualAttribute1} \\ \texttt{rdf:List}}] (EQA1) at (-8,4) {};
                \node[env, pattern=crosshatch, label={\text{ex:EqualAttribute2} \\ \texttt{rdf:List}}] (EQA2) at (-8,8.5) {};
                \node[env, pattern=crosshatch, label={\text{ex:Q} \\ \texttt{OM:Variable}}] (Q) at (-2,8) {};
                \node[env, pattern=crosshatch, label={\text{ex:VCavity} \\ \texttt{OM:Variable}}] (VC) at (-11,-4) {};
                \node[env, label=below:{\text{ex:FlowOfMass} \\ \texttt{VDI3682:Information}}] (FOM) at (-15,10) {};
                \node[env, label={\text{ex:Fibre} \\ \texttt{VDI3682:Product}}] (FIB) at (-19,12) {};
                \node[env, label={\text{ex:Resin} \\ \texttt{VDI3682:Product}}] (RES) at (-17,14) {};
                \node[env, label={\text{ex:FlowOfMass-A} \\ \texttt{DIN61360:DataElement}}] (FOM-A) at (-13.5,12.5) {};
                \node[env, label=below:{\text{ex:VolumeCMQ} \\ \texttt{DIN61360:TypeDescription} \\ \texttt{UNECE:CMQ}}] (VOL) at (-14,-7) {};
                \node[env, label={\text{ex:FlowRate} \\ \texttt{DIN61360:TypeDescription} \\ \texttt{UNECE:2J}}] (FLO) at (-2,12) {};
                \node[env, label=below:{\text{ex:RTM-Tool-B} \\ \texttt{VDI3682:TechnicalResource}}, draw=darkorange, fill=orange] (TRB) at (-3,-14) {};
                \node[env, label={\text{ex:CavityVolume-B} \\ \texttt{DIN61360:DataElement}},draw=darkorange,fill=orange] (HRB) at (-14,-12) {};
                \node[env, label=below:{\text{ex:VolumeLTR} \\ \texttt{DIN61360:TypeDescription}\\\texttt{UNECE:LTR}},draw=darkorange,fill=orange] (VLTR) at (-15,-14) {};
                \node[env, label=below:{\text{ex:InjectionHP} \\ \texttt{VDI3682:ProcessOperator}}, draw=darkorange, fill=orange] (POHP) at (-3,-11) {};
                \node[env, label=below:{\text{ex:FlowOfMassHP} \\ \texttt{VDI3682:Information}}, draw=darkorange,fill=orange] (IOHP) at (-3,-7.5) {};
                  \node[env, label=below:{\text{ex:FlowOfMassHP-B} \\ \texttt{DIN61360:DataElement}}, draw=darkorange, fill=orange] (FOMHP) at (-3,-5) {};
							]
			\end{scope}
			
			\begin{scope}[>={Stealth[black]},
				every node/.style={fill=white},
				every edge/.style={draw=black, font=\normalsize}]
				every label/.style={font=\scriptsize}
				\path [->] (TRA) edge[bend right=10] node [pos=0.6] {\texttt{DIN61360:hasDataElement}} (CVA); 
				\path [->] (PO) edge[bend right=60] node[pos=0.5] {\texttt{VDI3682:isAssignedTo}} (TRA);
		
              \path[->] (PO) edge[bend right=45] node[pos=0.4] {\texttt{ParX:hasInterdependency}} (AppT);
              \path [->] (App) edge[bend right=70] node[pos=0.6] {\texttt{OM:operator}} (Op);
               \path [->] (AppT) edge[bend right=70] node[pos=0.65] {\texttt{OM:operator}} (OpEq);
              \path [->] (App) edge[bend left=30] node[pos=0.5] {\texttt{OM:arguments}} (AL1);
               \path [->] (AppT) edge[bend left=30] node[pos=0.5] {\texttt{OM:arguments}} (EQA1);
              \path [->] (AL1) edge[bend right=30] node[pos=0.5] {\texttt{rdf:rest}} (AL2);
              \path [->] (AL2) edge [] node[pos=0.5] {\texttt{rdf:first}} (Q);
              \path [->] (AL1) edge [bend left=10] node[pos=0.3] {\texttt{rdf:first}} (VC);
              \path [->] (EQA1) edge [bend left=10] node[pos=0.5] {\texttt{rdf:first}} (t);
              \path [->] (EQA2) edge [bend left=10] node[pos=0.6] {\texttt{rdf:first}} (App);
              \path [->] (EQA1) edge [bend left=10] node[pos=0.7] {\texttt{rdf:rest}} (EQA2);
              \path [->] (CVA) edge[bend left=5] node[pos=0.5] {\texttt{ParX:isDataFor}} (VC);
              \path [thick, ->] (HRB) edge[bend right=50, darkorange, -latex, dashed] node[pos=0.5] {\texttt{ParX:isDataFor}} (VC);
              \path [->] (AL1) edge [bend left=10] node[pos=0.3] {\texttt{rdf:first}} (VC);
              \path [->] (FOM-A) edge [bend left=10] node[pos=0.5] {\texttt{ParX:isDataFor}} (Q);
              \path [->] (TDE) edge [bend right=30] node[pos=0.7] {\texttt{ParX:isDataFor}} (t);
              \path [->] (FOM) edge [bend left=10] node[pos=0.5] {\texttt{DIN61360:hasDataElement}} (FOM-A);
              \path [->] (PO) edge [bend left=10] node[pos=0.5] {\texttt{VDI3682:hasInput}} (FOM);
              \path [->] (PO) edge [bend left=10] node[pos=0.45] {\texttt{VDI3682:hasInput}} (FIB);
              \path [->] (PO) edge [bend left=5] node[pos=0.5] {\texttt{VDI3682:hasInput}} (RES);
              \path [->] (PO) edge [bend right=30] node[pos=0.4] {\texttt{VDI3682:hasOutput}} (OI);
            \path [->] (CVA) edge [bend right=30] node[pos=0.3] {\texttt{DIN61360:hasTypeDescription}} (VOL);
            \path [->] (FOM-A) edge [bend left=10] node[pos=0.5] {\texttt{DIN61360:hasTypeDescription}} (FLO);
            \path [->] (HRB) edge [bend right=35] node[pos=0.7] {\texttt{DIN61360:hasTypeDescription}} (VLTR);
            \path [->] (TRB) edge[bend right=20] node[pos=0.5] {\texttt{DIN61360:hasDataElement}} (HRB);
            \path [->] (OI) edge[bend right=20] node[pos=0.45] {\texttt{DIN61360:hasDataElement}} (TDE);
            \path [->] (Q) edge [bend right=5,  blue ] node[pos=0.5] {\texttt{ParX:expectsUnit}} (FLO);
            \path [->] (VC) edge [bend right=5, blue] node[pos=0.8] {\texttt{ParX:expectsUnit}} (VOL);
            \path [->] (POHP) edge[bend right=10] node[pos=0.4] {\texttt{VDI3682:isAssignedTo}} (TRB);
            \path [->] (POHP) edge [bend left=10] node[pos=0.5] {\texttt{VDI3682:hasInput}} (IOHP);
            \path [->] (IOHP) edge[bend right=20] node[pos=0.5] {\texttt{DIN61360:hasDataElement}} (FOMHP);
            \path [->] (FOMHP) edge [bend right=30,  darkorange, -latex, dashed] node[pos=0.5] {\texttt{ParX:isDataFor}} (Q);
            \path [->] (FOMHP) edge [bend right=40,  darkorange, -latex, dashed] node[pos=0.10] {\texttt{DIN61360:hasTypeDescription}} (FLO);
            \path[->] (POHP) edge[bend left=10, darkorange, -latex, dashed] node[pos=0.35] {\texttt{ParX:hasInterdependency}} (AppT);
			\end{scope}
		\end{tikzpicture}}
    \caption{Simplified excerpt of the modeled \ac{rtm} injection process step in the ontology, focusing on interdependencies, data element bindings, and unit semantics}
    \label{fig:Inconsistent}
\end{figure}

Subsequently, the model was enriched with relevant data, information, and interdependencies. 
For better readability, a simplified excerpt of the ontology is shown in \figref{fig:Inconsistent}.

The injection step of the \ac{rtm} process is modeled as a \texttt{VDI3682:ProcessOperator} that transforms resin and a dry fibre preform into a composite product using energy and process information such as temperature. 
While these transformation semantics and additional input/output states are part of the full ontology, they are omitted from the diagram for clarity. 
The figure focuses on illustrating the semantic structure used to describe parameter interdependencies and their data bindings. 
Specifically, the formula for calculating the fill time \( t_{\text{fill}} \), shown in \eqref{eq:filltime}, is represented using \textit{OpenMath-RDF} as an \texttt{OM:Application} node that divides the cavity volume~(\( V_{\text{cavity}} \)) by the flow rate ($Q $). In \figref{fig:Inconsistent}, the \textit{OpenMath-RDF} expression is illustrated by crossed circles.
Both variables are connected to respective \texttt{DIN61360:DataElement} instances via the \texttt{ParX:isDataFor} property. 
Expected units are declared using \texttt{ParX:expectsUnit}, while actual units are defined through type descriptions classified under UNECE unit codes~(e.g., \texttt{UNECE:CMQ} for volume). 

\begin{equation}
t_{\mathrm{fill}} = \frac{V_{\mathrm{cavity}}}{Q}
\label{eq:filltime}
\end{equation}

The \ac{sparql} \textit{context filter pattern} introduced in Section~\ref{subsec:relevantC} was integrated into the queries to ensure that only data elements relevant to the specific process context are returned. The injection step of the \ac{rtm} process serves as an example to demonstrate the necessity of this mechanism. As illustrated in \figref{fig:Inconsistent}, the variable $V_\text{cavity}$ is connected to multiple data elements originating from different tools used in different processes, as indicated by the orange (dashed) connection. Since the formula in \eqref{eq:filltime} is modeled in a general and reusable way, a query does not by itself distinguish between contextually relevant and irrelevant data. Without the application of the filtering pattern, \ac{sparql} queries may incorrectly include data elements from other processes, which can lead to invalid parameter bindings.

To validate the filtering mechanism, the retrieval of variables (including their associated data elements) linked to the process operator \texttt{ex:Injection} was tested both with and without applying the proposed filter function in SPARQL. 
This comparison illustrates how the filtering mechanism affects the selection of context-relevant data elements during query execution.

Table~\ref{tab:filterresults} summarizes the retrieved variable-to-data-element bindings in both cases for the fill time calculation. 
As shown, the unfiltered query yields additional data elements for the same variables (e.g., \texttt{ex:VCavity} linked to \texttt{ex:CavityVolume-B}), which originate from a different process colored in orange (high pressure injection \texttt{ex:InjectionHP}) and are thus not valid for the evaluation of \texttt{ex:Injection}. Similarly, the variable \texttt{ex:Q} is associated with both the correct input \texttt{ex:FlowOfMass-A} and the unrelated \texttt{ex:FlowOfMassHP-B}.

In contrast, the filtered query correctly restricts the results to only those data elements that are reachable via the structural context of the selected process operator, including outputs, inputs, and assigned technical resources. This confirms that the proposed pattern ensures semantic relevance and prevents cross-contextual contamination of variable bindings, which is critical for maintaining consistency in the evaluation of shared or reused interdependency descriptions.

\begin{table}[h]
\centering
\caption{SPARQL query response for fill time calculation in process step \texttt{ex:Injection} with / without context filter pattern}
\label{tab:filterresults}
\begin{tabular}{llc}
\toprule
\textbf{Variable} & \textbf{Data Element}     & \textbf{Filtered} \\
\midrule
\texttt{ex:Q}        & \texttt{ex:FlowOfMass\_A}     & \checkmark \\
\texttt{ex:Q}        & \texttt{ex:FlowOfMassHP\_B}         & X\\
\texttt{ex:VCavity}  & \texttt{ex:CavityVolume\_A}             & \checkmark \\
\texttt{ex:VCavity}  & \texttt{ex:CavityVolume\_B}            & X \\
\texttt{ex:t} & \texttt{ex:FillTime\_DE}                  & \checkmark \\
\bottomrule
\end{tabular}
\end{table}

The \textit{unit consistency verification} presented in Section~\ref{subsec:unitC} plays a critical role in ensuring that the fill time calculation is based on semantically valid input. 
In the example shown, \( V_{\mathrm{cavity}} \) is expected to be measured in the unit of volume \texttt{UNECE:CMQ}~($\text{cm}^3$), while the flow rate \( Q \) is expected in \texttt{UNECE:2J}~($\text{cm}^3/\text{s}$). 
These expectations are explicitly modeled using the \texttt{ParX:expectsUnit} property, which is visually represented in \figref{fig:Inconsistent} as blue edges pointing from each variable to the corresponding unit class. 

To validate the proposed unit consistency verification, a negative example (colored in orange) was intentionally introduced into the knowledge graph. 
The variable \texttt{ex:VCavity}, which semantically expects a volume unit of type \texttt{UNECE:CMQ} ($\text{cm}^3$), was associated with a data element (\texttt{ex:CavityVolume-B}) whose type description was classified as \texttt{UNECE:LTR} (litres) instead. 

This discrepancy was correctly detected using the \ac{sparql} query shown in Listing~\ref{lst:UnitConsistencyCheck} and highlights a common real-world issue in model-based engineering workflows.
In this case, the unit mismatch may result from the automated or semi-automated mapping of engineering data from datasheets, engineering tools or external sources.
If this information is imported into the semantic model without verifying unit consistency with the target variable's expectations in the interdependency formula (e.g., $\text{cm}^3$), semantic inconsistencies may arise unnoticed. 

As shown in Table~\ref{tab:unitResults}, the verification mechanism successfully identified the mismatch by comparing the expected and actual unit classifications. This confirms the mechanism's effectiveness in identifying critical modeling or mapping errors at design time.

\begin{table}[h]
\centering
\caption{SPARQL query response for unit inconsistencies}
\label{tab:unitResults}
\begin{tabular}{ll|ll}
\toprule
\textbf{Variable}  & \textbf{Expected Unit} & \textbf{Actual Unit} & \textbf{Data Element} \\
\midrule
\scriptsize{\texttt{ex:VCavity}}  & \scriptsize{\texttt{UNECE:CMQ}}  & \scriptsize{\texttt{UNECE:LTR}} & \scriptsize{\texttt{ex:CavityVolume-B} }\\
\bottomrule
\end{tabular}

\end{table}

In addition to correct unit usage, the completeness of data assignments is essential for evaluating interdependencies. 
To validate the proposed \textit{data availability verification}, an additional process description centered around a new process operator \texttt{ex:InjectionT} was inserted into the knowledge graph.

The process description follows the same structure as the previously modeled injection steps but intentionally omits the provision of a data element for the variable \( V_{\mathrm{cavity}} \). 
Specifically, the technical resource assigned to \texttt{ex:InjectionT} does not offer a \texttt{DINEN61360:DataElement} that could be connected to 
\( V_{\mathrm{cavity}} \) via the \texttt{ParX:isDataFor} property. The same interdependency formula graph for fill time calculation is reused.
While the \texttt{ex:InjectionT} process and its associated entities were inserted into the knowledge graph for verification purposes, they are not depicted in \figref{fig:Inconsistent} to maintain figure clarity and readability.

The data availability verification query (see Listing~\ref{lst:ComputabilityyCheck}) was executed, incorporating the context-filter pattern introduced in Section~\ref{subsec:relevantC}.

As summarized in Table~\ref{tab:datacheck}, the mechanism successfully detected that the variable \( V_{\mathrm{cavity}} \) lacks a context-relevant data element for the process \texttt{ex:InjectionT}. 
Thus, the mechanism correctly identified that the interdependency effect on $t_\text{fill}$ cannot be evaluated for this process due to incomplete knowledge modeling.

This demonstrates that the proposed verification approach identifies modeling gaps at design time, thereby supporting the creation of machine-interpretable and computationally valid process descriptions.

\begin{table}[h]
\centering
\caption{SPARQL query response for data availability}
\label{tab:datacheck}
\begin{tabular}{ll}
\toprule
\textbf{Missing Data for Variable} & \textbf{Process Context} \\
\midrule
\texttt{ex:VCavity} & \texttt{ex:InjectionT} \\
\bottomrule
\end{tabular}
\end{table}

\section{Conclusion and Future Work} \label{sec:conclusion}
This paper presents a set of verification mechanisms to ensure semantic consistency in ontology-based process models that integrate formal parameter interdependencies. 
These mechanisms are based on an aligned ontology design that combines domain-specific modeling patterns based on industrial standards with a mathematical representation of parameter interdependencies.

The proposed approach addresses three typical challenges in the context of interdependency-based knowledge models. First, it ensures the correct retrieval of results during query execution. Second, it supports the detection and avoidance of modeling errors that may compromise the evaluability or correctness of such calculations. This is achieved through (i) the contextual filtering of data elements using a \ac{sparql}-based query pattern, (ii) the validation of unit compatibility based on expected-unit annotations, and (iii) the verification of data availability through \ac{sparql}-based traversal of \textit{OpenMath-RDF} expressions.

By embedding these verification mechanisms into the ontology layer, the approach strengthens the consistency and, thereby, the reliability and reusability of formalized interdependencies. 
This is particularly relevant in scenarios involving reusable equations, heterogeneous data sources, or design-time assessments of \ac{cps} behavior.
It enables the detection of modeling errors early and therefore ensures a reliable evaluation of process configurations using ontology-based interdependency descriptions of manufacturing processes. 
As demonstrated through the \ac{rtm} process use case, this integration supports precise modeling and verification of process-specific interdependencies and facilitates the derivation of process outputs or performance indicators from structured domain knowledge.

Despite these advantages, interacting with complex knowledge graphs remains a significant challenge for many users, especially those unfamiliar with ontology engineering or semantic technologies. 
This highlights the need for assistance systems that are closely tailored to the requirements and skills of end users. 
A promising direction has been explored by \citet{Reif.2024c}, who investigate the use of large language models (LLMs) to enable intuitive interaction with knowledge graphs. 
Integrating and extending such approaches within the presented framework could further improve usability and accessibility, particularly for domain experts without technical modeling expertise.
Future work will focus on knowledge-based assistance systems that allow interdependencies to be directly calculated based on the knowledge graph and stored data.

A limitation of the current approach lies in the prerequisite that parameter interdependencies must be explicitly available in formalized mathematical form. In some engineering scenarios, relations are not always fully known or easily specifiable. 
Machine learning methods may play a valuable role in discovering and approximating such unknown interdependencies. 
Once identified, these relations may be transformed into formal representations and integrated into the knowledge model, thereby closing knowledge gaps and expanding the model’s applicability. 

Additional validation will be pursued to assess the practicality, scalability, and robustness of the approach. 
Future investigations will examine whether the proposed verification mechanisms remain effective when applied to large-scale knowledge graphs with extensive interdependencies. 
This will help to evaluate potential performance bottlenecks in query execution and reasoning tasks, as well as the general applicability of the approach in complex manufacturing environments.

\section*{Acknowledgment}
This research paper [projects \textit{LaiLa}, \textit{iMOD}] is funded by \textit{dtec.bw – Digitalization and Technology Research Center of the Bundeswehr} which we gratefully acknowledge. \textit{dtec.bw} is funded by the \textit{European Union – NextGenerationEU}.
\interlinepenalty=10000
\bibliographystyle{IEEEtranN}
\bibliography{./bibliography/References.bib} 
\end{document}